\makeatletter
\@namedef{ver@everyshi.sty}{}
\makeatother
\documentclass[final]{cvpr}
\usepackage{tikz}
\usepackage{times}
\usepackage{epsfig}
\usepackage{graphicx}
\usepackage{amsmath}
\usepackage{amssymb}
\usepackage[acronym]{glossaries}
\usepackage{subcaption}
\usepackage{makecell}
\usepackage{arydshln}

\usepackage[pagebackref=true,breaklinks=true,colorlinks,bookmarks=false]{hyperref}

\newacronym{cnn}{CNN}{Convolutional Neural Network}
\newacronym{roi}{ROI}{Region Of Interest}
\newacronym{ui}{UI}{User Interface}
\graphicspath{{Images/}}

\begin{document}

\title{An Experience-based Direct Generation approach to Automatic Image Cropping}

\author{Casper Christensen\\
Gracenote, A Nielsen Company\\
\and
Aneesh Vartakavi\\
Gracenote, A Nielsen Company\\
{\tt\small aneesh.vartakavi@nielsen.com}
}

\maketitle

\begin{abstract}
Automatic Image Cropping is a challenging task with many practical downstream applications. The task is often divided into sub-problems - generating cropping candidates, finding the visually important regions, and determining aesthetics to select the most appealing candidate. Prior approaches model one or more of these sub-problems separately, and often combine them sequentially. 
We propose a novel convolutional neural network (CNN) based method to crop images directly, without explicitly modeling image aesthetics, evaluating multiple crop candidates, or detecting visually salient regions. Our model is trained on a large dataset of images cropped by experienced editors and can simultaneously predict bounding boxes for multiple fixed aspect ratios.
We consider the aspect ratio of the cropped image to be a critical factor that influences aesthetics. Prior approaches for automatic image cropping, did not enforce the aspect ratio of the outputs, likely due to a lack of datasets for this task. We, therefore, benchmark our method on public datasets for two related tasks - first, aesthetic image cropping without regard to aspect ratio, and second, thumbnail generation that requires fixed aspect ratio outputs, but where aesthetics are not crucial. We show that our strategy is competitive with or performs better than existing methods in both these tasks. Furthermore, our one-stage model is easier to train and significantly faster than existing two-stage or end-to-end methods for inference. We present a qualitative evaluation study, and find that our model is able to generalize to diverse images from unseen datasets and often retains compositional properties of the original images after cropping. We also find that the model can generate crops with better aesthetics than the ground truth in the MIRThumb dataset for image thumbnail generation with no fine tuning. Our results demonstrate that explicitly modeling image aesthetics or visual attention regions is not necessarily required to build a competitive image cropping algorithm. 

\end{abstract}


\section{Introduction}
\label{sec:introduction}

With the proliferation of devices like smartphones, smart televisions, and tablets, imagery in different aspect ratios is necessary for a user interface to comply with responsive web design standards. These images are often manually cropped, which can be very laborious to perform for a large number of images. Automatic Image Cropping, therefore, has great practical significance for large catalogs of images. An effective aesthetic cropping algorithm could be helpful to industries and applications that store and display large amounts of media, such as social networks or image sharing platforms, image galleries, surveillance systems, photography and graphic design software.

Image cropping is often performed to highlight visual attention regions discarding unwanted regions in the process. Alternatively or in conjunction, cropping can be performed to improve or maintain the aesthetics of an image. Experienced users, including trained photographers, may use composition concepts such as the rule of thirds or the golden ratio to maximize aesthetics while deciding how to crop images. The aspect ratio of the final cropped image is also essential when performing this task, as it affects the aesthetics and the framing of the image. For example, selecting a portrait crop from a landscape image with multiple subjects can only include a subset of them, and the final crop should not include any partially cropped faces for aesthetic reasons. Advances in image cropping methods could therefore inform and guide research in visual perception and aesthetics.

The detection of visual attention regions in images has been an active area of research for some time \cite{ullman1988structural}. Attention-based automatic cropping approaches build on it by drawing bounding boxes around the image's salient regions, assuming that the best crop should include the salient region. Assessment of image aesthetics is also an active research area, starting with low-level rules and features, which are difficult to formulate and do not generalize, to recent deep learning approaches \cite{deng2017image}. 
The aspect ratio of an image is also essential to the perceived aesthetics, recognized by some recent aesthetics assessment approaches \cite{wang2019aspect,chen2020adaptive}. However, it is rarely mentioned as a requirement or concern in prior approaches to automatic image cropping.  Although some techniques can output bounding boxes in different aspect ratios \cite{tu2019image,zeng2019grid}, they do so by evaluating multiple candidates and are therefore inefficient. Image cropping is a typical first stage for thumbnail generation approaches, which try to create smaller representations of images. These strategies often create thumbnails in fixed aspect ratios but usually do not consider image aesthetics \cite{deng2017image,chen2018cropnet}.

Early approaches to automatic image cropping tended to focus on either the aesthetics or the visual attention regions. More recent solutions try to incorporate both by modeling the cropping process in two stages. First, they determine a visual attention region or a \acrfull{roi}, and then, they draw a bounding box to maximize aesthetics. This two-stage approach has some disadvantages: when the image has no salient regions \cite{lu2019aesthetic,tu2019image}, or when it has multiple salient subjects, some of which may need to be excluded for aesthetic reasons \cite{tu2019image, lu2019end}.

We believe that a single-stage approach that implicitly models the image's aesthetics and attention regions can overcome some of the drawbacks of existing image cropping techniques. Our proposed model is less susceptible to failure cases that occur when attention or aesthetics are modeled explicitly, such as, when no salient region is found or when the ground truth for aesthetic assessment is ambiguous due to neutral image aesthetics \cite{deng2017image}.  
We evaluate several common CNN architectures in a transfer learning framework and find that a WideResNet50-2\mbox{\cite{zagoruyko2016wide}} backend achieves the best overall performance on our dataset with an IoU of $0.867$. This model is more lightweight and efficient than two-stage approaches and is simpler to train. Without any model optimization or pruning, our model can process over $600$ images/sec, or over $3000$ crops/sec as each image is cropped in $5$ aspect ratios on a single Nvidia Tesla V100 GPU during inference. This is significantly faster than existing approaches \cite{chen2018cropnet, lu2019end, lu2019listwise, lu2019aesthetic,lu2020learning}. 


To the best of our knowledge, this work is the first attempt at addressing the problem of image cropping directly, without explicitly modeling visual attention or aesthetics.Due to a lack of public datasets to support our approach, we train our CNN-based model using a large internal dataset of images cropped by experienced editors in fixed aspect ratios, who simultaneously maintain image aesthetics and important image content. We propose an efficient architecture that predicts bounding boxes for multiple aspect ratios simultaneously, without evaluating multiple crop candidates. Prior approaches for image cropping did not enforce the aspect ratio of their outputs. We, therefore, benchmark our task on datasets for two related tasks - FCDB \cite{chen2017quantitative} for aesthetic image cropping without regard to aspect ratio, and MIR-Thumb\cite{chen2018cropnet} for thumbnail generation in fixed aspect ratios where aesthetics are not crucial. Our model with a WideResNet50-2 backend, modified to generate outputs in any aspect ratio, is competitive with and more efficient than existing approaches on FCDB, achieving an IoU of 0.692. We also achieve state-of-the-art performance on the MIR-Thumb dataset at an IoU of 0.741 with no fine-tuning. This demonstrates that explicitly modeling aesthetics or attention regions is not strictly required for accurate and efficient image cropping. Finally, we include a qualitative evaluation, where we investigate the generalization ability of the model on the FCDB and MIRThumb datasets without fine tuning. Where also observe that the model can generate more aesthetic crops on MIR-Thumb than the original ground truth. This finding highlights some challenges in the objective evaluation of image cropping systems, such as the reliance on crowd-sourced workers to gather ground truth and using a single reference for the IoU metric when several equally good crops may exist.


In summary:
\begin{itemize}
\item We are the first work to attempt aesthetic image cropping directly and show that explicitly modeling visual attention or image aesthetics is not necessary to build a competitive image cropping algorithm.

\item We propose a simple architecture, with no bells and whistles that is easier to train compared to recent state-of-the-art approaches, such as separated network branches for bounding box prediction \cite{chen2018cropnet}, ROI-aware pooling operations \cite{chen2018cropnet, lu2019end, lu2019listwise}, human-defined composition patterns \cite{tu2019image}, and custom loss functions \cite{lu2019end,tu2019image}. 

\item Our proposed single-stage model is efficient and able to output bounding boxes of multiple fixed aspect ratios, without evaluating multiple candidates, which is novel for aesthetic aware image cropping approaches.


\end{itemize}

\section{Related Work}

Prior approaches to solve the automatic image cropping problem can be distinguished by how the cropping candidates are initially determined and how they are evaluated to get the final crop. The task of selecting cropping candidates is generally solved by a few different approaches:
\begin{itemize}
    \item \emph{Sliding-Judging} - These techniques generate a large number of candidates by moving windows of varying sizes and aspect ratios over the original image, each of which is then evaluated against some criterion such as image aesthetics or attention regions to find the best candidate \cite{sun2013scale,zeng2019grid,nishiyama2009sensation,fang2014automatic}. These strategies are generally computationally inefficient as the search space spans the entire image \cite{lu2019end, wang2017deep}. Some authors have developed strategies to mitigate this by exploiting properties such as local redundancy \cite{zeng2019grid} or by eliminating candidates that do not encompass the entire region of interest \cite{yan2013learning}. Other authors suggest more efficient solutions that evaluate fewer candidates, but without regard to aesthetics \cite{chen2016automatic}.

    \item \emph{Determining-adjusting} - These methods try to first determine an \acrshort{roi} in the image. They then generate many candidates around that region by adjusting the position, height, or aspect ratio of the bounding boxes, and evaluate each of them to find the best cropping candidate \cite{wang2017deep,wang2018deep}. They are more efficient than sliding-judging approaches because they generate fewer candidates, but they struggle when no \acrshort{roi} is found \cite{lu2019aesthetic, lu2019end}. 
    
    \item \emph{Finding-generating} - These methods aim to predict a single crop region by calculating a bounding box that includes the visual attention region in the image. This is then fed into a regression network that predicts the optimal bounding box \cite{lu2019aesthetic,lu2019end}. These strategies are efficient because they generate a single candidate, instead of generating and evaluating multiple candidates as in determining-adjusting approaches. However, these methods also struggle when no \acrshort{roi} is found \cite{lu2019aesthetic, lu2019end}.

\end{itemize}

Once the candidates are generated, prior approaches evaluate them in a few different ways:
\begin{itemize}
    \item \emph{Saliency} or attention-based methods assume that the best crop will generally contain the most salient regions. The techniques for finding the salient regions range from signal processing \cite{itti1998model} to deep learning methods \cite{vig2014large, liu2015predicting, wang2019salient}. \emph{Determining-adjusting} approaches often use these methods to find an \acrshort{roi} \cite{wang2017deep,wang2018deep}. Other saliency-based cropping methods include Ardizzone et al. \cite{ardizzone2013saliency}, Ciocca et al. \cite{ciocca2007self}, and Sun and Ling \cite{sun2013scale}.
    
    \item \emph{Aesthetic evaluation} methods try to quantify and score images or crop candidates based on their aesthetic qualities. A comprehensive review of these methods is presented by Deng et al. \cite{deng2017image}. Aesthetic image cropping algorithms sometimes use features inspired by composition rules such as the rule of thirds and visual balance \cite{khuan2016aaics,tu2019image,yan2013learning}. Datasets such as AVA \cite{murray2012ava} enable learning aesthetics using deep learning methods \cite{mai2016composition}. Other aesthetics-based image cropping approaches include Nishiyama et al. \cite{nishiyama2009sensation}, Zhang et al. \cite{zhang2012probabilistic}, and Chen et al. \cite{chen2017learning}.
    
    \item \emph{Fusion} methods try to combine attention and aesthetic methods in two stages and harness the advantages of both. Some approaches use a determining-adjusting strategy by first predicting the attention region, then generating a small number of candidates around it, and finally selecting the one with the best aesthetic evaluation score \cite{wang2017deep,wang2018deep}. Finding-generation strategies try to regress the bounding box after detecting salient regions in an image \cite{lu2019aesthetic,lu2019end}. Other fusion approaches include Tu et al. \cite{tu2019image} , Guo et al. \cite{guo2018automatic}, and Li et al. \cite{li2018a2}.
    
    \item \emph{Experience-based} methods try to predict a bounding box using a dataset of images cropped by humans. Prior methods that follow this strategy design handcrafted features such as sharpness and color distance that are then used for regressing the bounding boxes \cite{yan2013learning,yan2015change}.

\end{itemize}

Some image cropping methods do not easily fit into this framework: a reinforcement learning framework \cite{li2018a2}, rank-based evaluation metrics on a densely annotated dataset \cite{zeng2019grid,zeng2019reliable}, weakly supervised learning \cite{lu2020weakly}, and rank-based learning approaches \cite{lu2019listwise,chen2017quantitative}. 

We propose an \emph{experience-based direct generation} strategy, which has not been attempted for aesthetic image cropping to the best of our knowledge. We propose the term \emph{direct generation} to represent methods that predict the bounding box directly from an input image, without the overhead of detecting visual attention regions or evaluating multiple cropping candidates. These methods do not suffer from the same drawbacks as \emph{finding-generating} approaches, such as when an \acrshort{roi} is absent, and are more efficient than \emph{sliding-judging} and \emph{determining-adjusting} methods. Our model is trained to directly predict the bounding boxes for different aspect ratios simultaneously, using a shared feature extractor for efficiency. We build a large internal dataset to train our experience-based approach with no handcrafted features, overcoming the limitations that restricted other methods \cite{lu2019aesthetic,lu2019end,guo2018automatic,wei2018good}.

There are some use cases where efficiency is not as important and evaluating multiple candidates may be desired, for example when presenting multiple candidates to a user and allowing them to pick the best candidate based on their preferences. However, in this work, we continue a trend in prior work that focuses on applications that benefit from reducing the number of evaluated candidates for efficiency reasons.

Image cropping is related to thumbnail generation, which aims to create smaller representative versions of the original images by preserving the most useful content from the original image and discarding the background. In contrast, image cropping approaches try to create new images, balancing aesthetic quality while including visually salient regions. Our approach is similar to some recent thumbnail generation approaches such as FastAT \cite{esmaeili2017fast}, and CropNet \cite{chen2018cropnet} in that they predict an output without generating multiple candidates. CropNet uses a similar strategy of a shared feature extractor and dedicated branches to predict multiple bounding boxes of fixed aspect ratios but follows a different strategy of predicting bounding boxes. CropNet is trained on MIR-Thumb, a smaller crowd-sourced dataset annotated by non-experienced workers, in contrast with our larger dataset annotated by trained experts who also pay attention to image aesthetics. In Section \ref{sec:experiments}, we benchmark our approach on the MIR-Thumb test set and achieve state-of-the-art performance with no fine-tuning. We also demonstrate that our algorithm can produce more aesthetically pleasing images and display some examples.

\section{Our Approach}
\subsection{Dataset}

\begin{figure}[ht]
    \begin{subfigure}[b]{.45\textwidth}
    \centering
    \includegraphics[width=\textwidth]{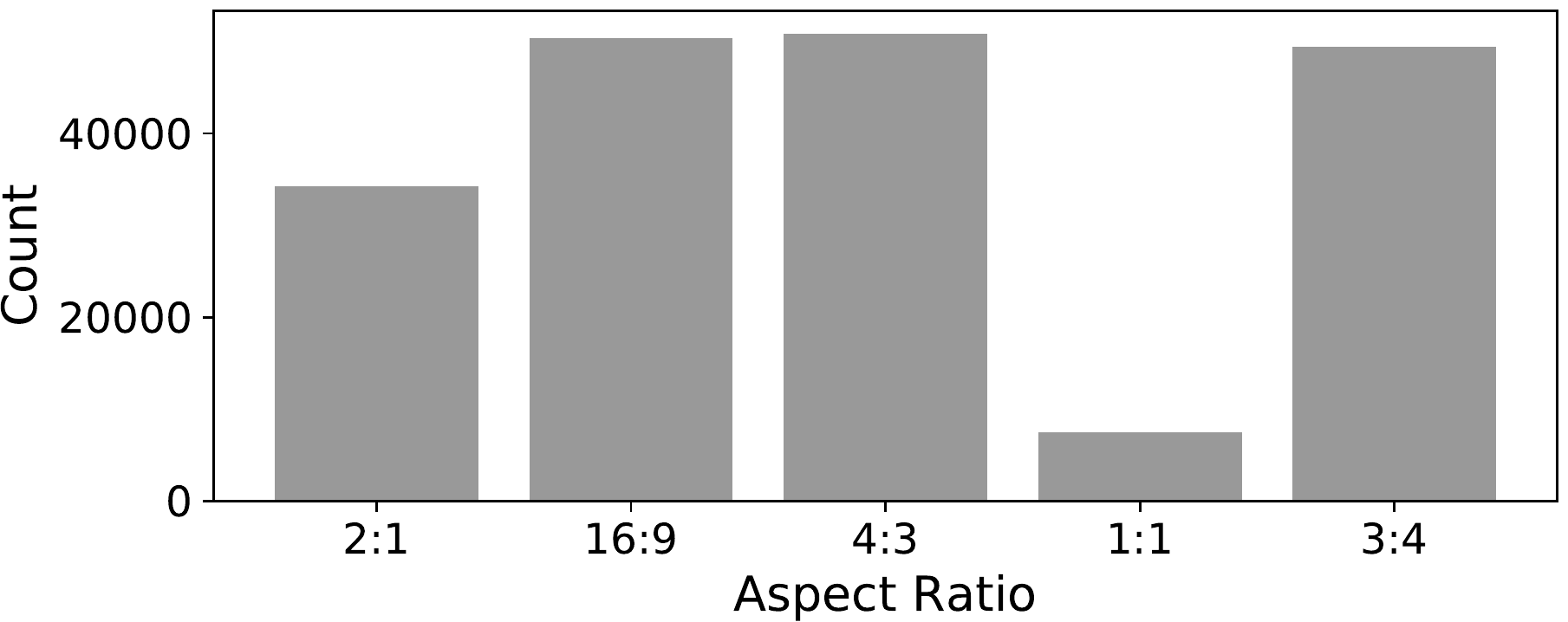}
    \caption{Distribution of aspect ratios of cropped images}
    \label{fig:master_aspect_ratio_dist}
    \end{subfigure}
    \hfill
    
    \begin{subfigure}[b]{.45\textwidth}
    \centering
    \includegraphics[width=\textwidth]{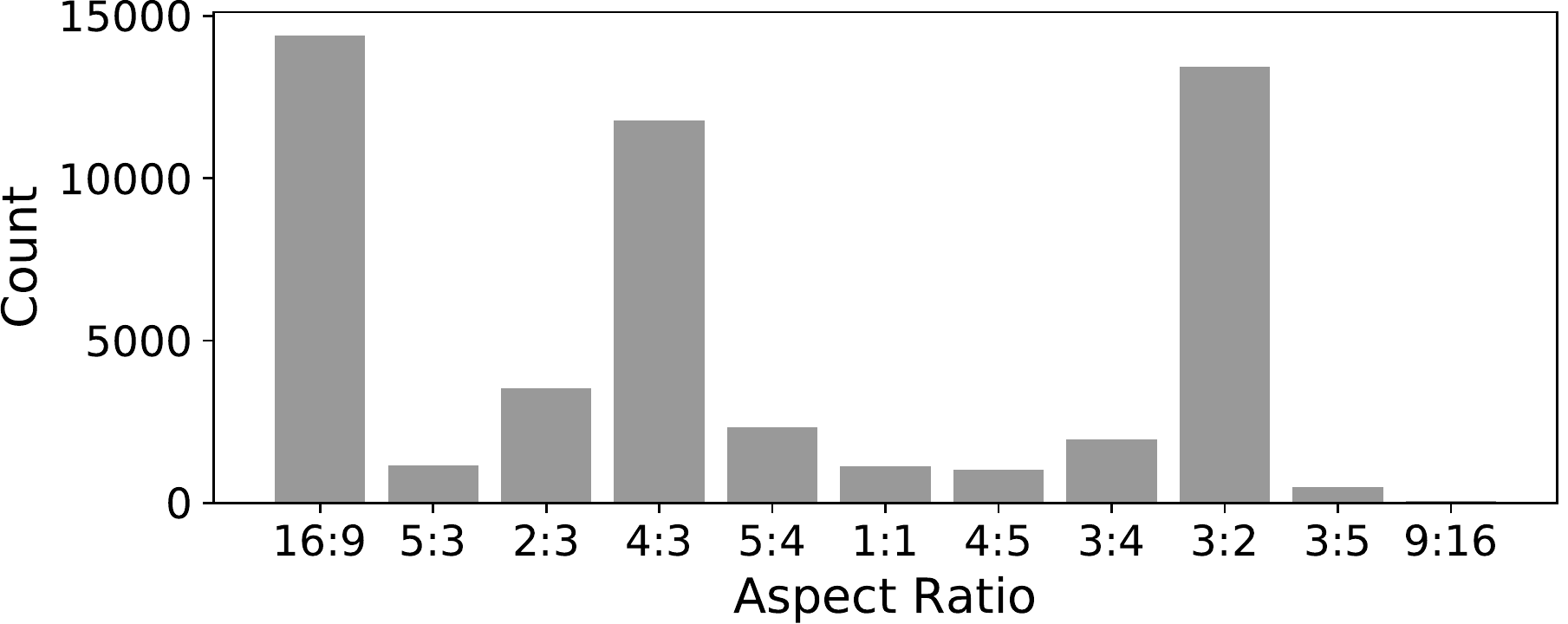}
    \caption{Distribution of aspect ratios of original images}
    \label{fig:raw_aspect_ratio_dist}
    \end{subfigure}
    \hfill
    \begin{subfigure}[b]{.45\textwidth}
    \centering
    \includegraphics[width=\columnwidth]{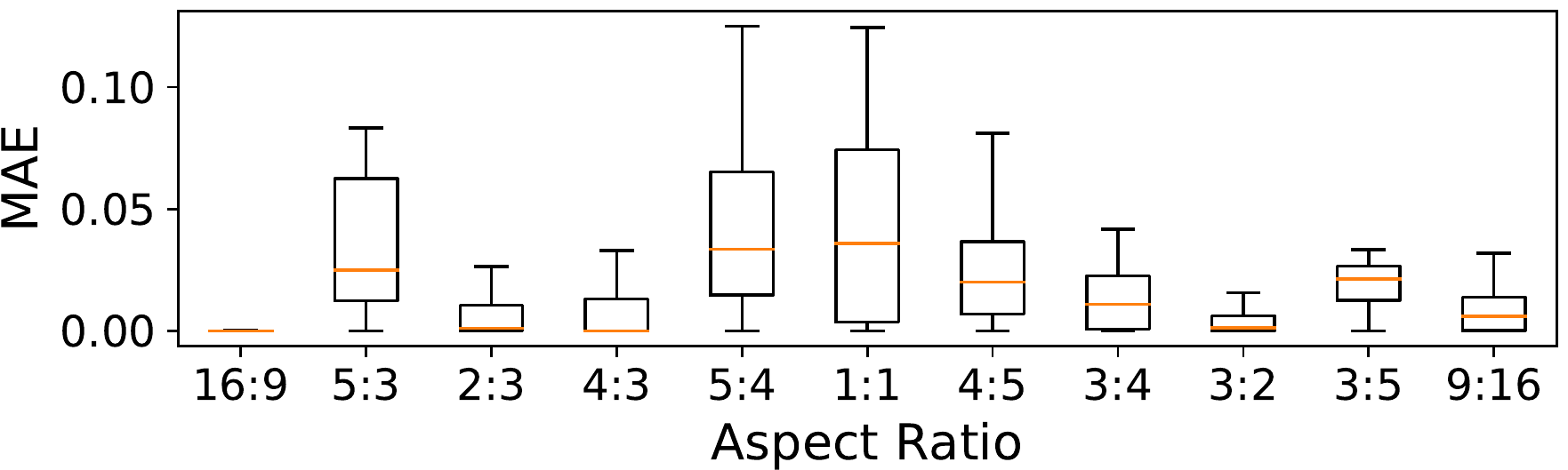}
    \caption{Mean Absolute Error (MAE) between the aspect ratio of the original image and the closest aspect ratio}
    \label{fig:raw_aspect_errors}
    
    \end{subfigure}
\caption{Aspect Ratio Distributions}
\label{fig:dataset_diversity}
\end{figure}

Prior approaches \cite{lu2019aesthetic,lu2019end,guo2018automatic,wei2018good} often cite the lack of large datasets for effective image cropping and design workarounds to overcome this limitation. We were unable to find existing datasets that support our experience-based direct generation approach to image cropping with strict aspect ratio requirements. We therefore collect an internal dataset of about $51,000$ images for this study, serving as iconic imagery for TV programs and movies. The images usually include the lead characters along with a background that conveys context relevant to the program. Each image was manually cropped in up to 5 aspect ratios (16:9, 4:3, 2:1, 3:4, and 1:1) by a large group of experienced editors who were asked to retain important image content, preserve the aesthetics, and adhere to strict aspect ratio requirements. Unlike some datasets such as FCDB, we did not rate or discard images based on their aesthetics in an effort to mitigate subjective bias. Some prior datasets suggested bounding boxes for their workers to rank or annotate, citing efficiency reasons \cite{wei2018good,chen2017quantitative}. In contrast, we allowed our editors to crop the images directly to avoid bias. Only a single editor was allowed to crop a given image in one aspect ratio, and no ranking or rating information was collected. We present some examples of these images in Section \ref{sec:qualitative}. Of all the images in the resulting dataset, not every image was cropped in every aspect ratio, as can be seen in Figure \ref{fig:master_aspect_ratio_dist}. The dataset is also diverse in the aspect ratios of the original images, as illustrated in Figure \ref{fig:raw_aspect_ratio_dist}. A successful model would have to generalize to input images of many sizes. We consider other common aspect ratios in addition to those mentioned above for this visualization. We also compute the mean absolute error of each image to the closest aspect ratio in Figure \ref{fig:raw_aspect_errors}, similar to the analysis performed by Celona et al. \cite{celona2019autocropping}.

\begin{figure*}[ht]
\begin{subfigure}[b]{\textwidth}
\centering
\tikzset{every picture/.style={line width=0.75pt}} 
\begin{tikzpicture}[x=0.75pt,y=0.75pt,yscale=-0.85,xscale=0.85]

\draw  [fill={rgb, 255:red, 0; green, 0; blue, 0 }  ,fill opacity=0.8 ] (18.39,40) -- (67.39,40) -- (52.36,82) -- (3.36,82) -- cycle ;
\draw  [color={rgb, 255:red, 0; green, 0; blue, 0 }  ,draw opacity=0.5 ][fill={rgb, 255:red, 155; green, 155; blue, 155 }  ,fill opacity=0.5 ] (328,32) -- (348,32) -- (348,52) -- (328,52) -- cycle ;
\draw  [color={rgb, 255:red, 0; green, 0; blue, 0 }  ,draw opacity=0.5 ][fill={rgb, 255:red, 155; green, 155; blue, 155 }  ,fill opacity=0.5 ] (338,42) -- (358,42) -- (358,62) -- (338,62) -- cycle ;
\draw  [color={rgb, 255:red, 0; green, 0; blue, 0 }  ,draw opacity=0.5 ][fill={rgb, 255:red, 155; green, 155; blue, 155 }  ,fill opacity=0.5 ] (348,52) -- (368,52) -- (368,72) -- (348,72) -- cycle ;
\draw  [color={rgb, 255:red, 0; green, 0; blue, 0 }  ,draw opacity=0.5 ][fill={rgb, 255:red, 155; green, 155; blue, 155 }  ,fill opacity=0.5 ] (358,62) -- (378,62) -- (378,82) -- (358,82) -- cycle ;
\draw  [color={rgb, 255:red, 0; green, 0; blue, 0 }  ,draw opacity=0.5 ][fill={rgb, 255:red, 128; green, 128; blue, 128 }  ,fill opacity=0.5 ] (138,42) -- (188,42) -- (188,92) -- (138,92) -- cycle ;
\draw  [color={rgb, 255:red, 0; green, 0; blue, 0 }  ,draw opacity=0.5 ][fill={rgb, 255:red, 128; green, 128; blue, 128 }  ,fill opacity=0.5 ] (128,32) -- (178,32) -- (178,82) -- (128,82) -- cycle ;
\draw  [color={rgb, 255:red, 0; green, 0; blue, 0 }  ,draw opacity=0.5 ][fill={rgb, 255:red, 128; green, 128; blue, 128 }  ,fill opacity=0.5 ] (118,22) -- (168,22) -- (168,72) -- (118,72) -- cycle ;
\draw  [color={rgb, 255:red, 0; green, 0; blue, 0 }  ,draw opacity=0.5 ][fill={rgb, 255:red, 128; green, 128; blue, 128 }  ,fill opacity=0.5 ] (228,32) -- (268,32) -- (268,72) -- (228,72) -- cycle ;
\draw  [dash pattern={on 4.5pt off 4.5pt}] (108,12) -- (388,12) -- (388,122) -- (108,122) -- cycle ;
\draw  [color={rgb, 255:red, 0; green, 0; blue, 0 }  ,draw opacity=0.5 ][fill={rgb, 255:red, 245; green, 166; blue, 35 }  ,fill opacity=1 ] (138,42) -- (158,42) -- (158,62) -- (138,62) -- cycle ;
\draw  [color={rgb, 255:red, 0; green, 0; blue, 0 }  ,draw opacity=0.5 ][fill={rgb, 255:red, 128; green, 128; blue, 128 }  ,fill opacity=0.5 ] (248,52) -- (288,52) -- (288,92) -- (248,92) -- cycle ;
\draw  [color={rgb, 255:red, 0; green, 0; blue, 0 }  ,draw opacity=0.5 ][fill={rgb, 255:red, 128; green, 128; blue, 128 }  ,fill opacity=0.5 ] (238,42) -- (278,42) -- (278,82) -- (238,82) -- cycle ;
\draw  [color={rgb, 255:red, 0; green, 0; blue, 0 }  ,draw opacity=0.5 ][fill={rgb, 255:red, 245; green, 166; blue, 35 }  ,fill opacity=1 ] (248,52) -- (258,52) -- (258,62) -- (248,62) -- cycle ;
\draw [color={rgb, 255:red, 0; green, 0; blue, 0 }  ,draw opacity=0.5 ]   (158,42) -- (248,52) ;
\draw [color={rgb, 255:red, 0; green, 0; blue, 0 }  ,draw opacity=0.5 ]   (158,62) -- (248,62) ;

\draw   (398,59.5) -- (416,59.5) -- (416,57) -- (428,62) -- (416,67) -- (416,64.5) -- (398,64.5) -- cycle ;
\draw  [fill={rgb, 255:red, 109; green, 151; blue, 67 }  ,fill opacity=0.5 ] (438,12) -- (448,12) -- (448,122) -- (438,122) -- cycle ;
\draw  [fill={rgb, 255:red, 74; green, 144; blue, 226 }  ,fill opacity=0.5 ] (500,12.21) -- (600,12.21) -- (600,30.37) -- (500,30.37) -- cycle ;

\draw  [line width=0.75]  (461,20.5) -- (479,20.5) -- (479,18) -- (491,23) -- (479,28) -- (479,25.5) -- (461,25.5) -- cycle ;
\draw   (611,19.5) -- (629,19.5) -- (629,17) -- (641,22) -- (629,27) -- (629,24.5) -- (611,24.5) -- cycle ;
\draw  [fill={rgb, 255:red, 74; green, 144; blue, 226 }  ,fill opacity=0.5 ] (500,52.21) -- (600,52.21) -- (600,70.37) -- (500,70.37) -- cycle ;
\draw  [line width=0.75]  (461,60.5) -- (478,60.5) -- (478,58) -- (490,63) -- (478,68) -- (478,65.5) -- (461,65.5) -- cycle ;
\draw   (610,59.5) -- (628,59.5) -- (628,57) -- (640,62) -- (628,67) -- (628,64.5) -- (610,64.5) -- cycle ;

\draw  [fill={rgb, 255:red, 74; green, 144; blue, 226 }  ,fill opacity=0.5 ] (500,102) -- (600,102) -- (600,120.16) -- (500,120.16) -- cycle ;
\draw  [line width=0.75]  (461,110.29) -- (479,110.29) -- (479,107.79) -- (491,112.79) -- (479,117.79) -- (479,115.29) -- (461,115.29) -- cycle ;
\draw   (613,106.5) -- (631,106.5) -- (631,104) -- (643,109) -- (631,114) -- (631,111.5) -- (613,111.5) -- cycle ;
\draw   (72,59.5) -- (90,59.5) -- (90,57) -- (102,62) -- (90,67) -- (90,64.5) -- (72,64.5) -- cycle ;
\draw  [fill={rgb, 255:red, 255; green, 255; blue, 255 }  ,fill opacity=1 ] (16.68,43.89) -- (65.12,43.89) -- (53.59,76.14) -- (5.15,76.14) -- cycle ;

\draw (225,133) node [anchor=north west][inner sep=0.75pt]   [align=center] {Feature\\
Extractor
};
\draw (420,133) node [anchor=north west][inner sep=0.75pt]   [align=center] {Feature\\Vector};
\draw (518,133) node [anchor=north west][inner sep=0.75pt]   [align=center] {
Regression\\
Heads};
\draw (533,15) node [anchor=north west][inner sep=0.75pt]  [font=\small] [align=left] {16:9};
\draw (649,14.4) node [anchor=north west][inner sep=0.75pt]  [font=\small]  {$( 0.01,\ 0.23,\ 0.99,\ 0.78)$};
\draw (648,54.4) node [anchor=north west][inner sep=0.75pt]  [font=\small]  {$( 0.02,\ 0.01,\ 1.00,\ 0.99)$};
\draw (537,55) node [anchor=north west][inner sep=0.75pt]  [font=\small] [align=left] {1:1};
\draw (651,101.4) node [anchor=north west][inner sep=0.75pt]  [font=\small]  {$( 0.13,\ 0.01,\ 0.87,\ 0.99)$};
\draw (541,105) node [anchor=north west][inner sep=0.75pt]  [font=\small] [align=left] {3:4};
\draw (548,76) node [anchor=north west][inner sep=0.75pt]  [font=\Large,rotate=-90] [align=left] {...};
\draw (301,60) node [anchor=north west][inner sep=0.75pt]  [font=\Large] [align=left] {...};
\draw (139,94) node [anchor=north west][inner sep=0.75pt]   [align=left] {{\tiny Convolution}};
\draw (194,69) node [anchor=north west][inner sep=0.75pt]   [align=left] {{\tiny Max Pool}};
\draw (245,94) node [anchor=north west][inner sep=0.75pt]   [align=left] {{\tiny Convolution}};
\draw (10,98) node [anchor=north west][inner sep=0.75pt]   [align=center] {Input\\
Image
};

\end{tikzpicture}
\caption{Model Architecture}
\label{fig:model_architecture}
\end{subfigure}
\hfill

\begin{subfigure}[b]{\textwidth}
\centering
\tikzset{every picture/.style={line width=0.75pt}} 

\begin{tikzpicture}[x=0.75pt,y=0.75pt,yscale=-0.85,xscale=0.9]

\draw  [fill={rgb, 255:red, 74; green, 144; blue, 226 }  ,fill opacity=0.5 ] (123,30) -- (153,30) -- (153,190) -- (123,190) -- cycle ;
\draw  [fill={rgb, 255:red, 255; green, 255; blue, 255 }  ,fill opacity=0.6 ] (128,44.57) .. controls (128,39.05) and (132.48,34.57) .. (138,34.57) .. controls (143.52,34.57) and (148,39.05) .. (148,44.57) .. controls (148,50.09) and (143.52,54.57) .. (138,54.57) .. controls (132.48,54.57) and (128,50.09) .. (128,44.57) -- cycle ;
\draw  [fill={rgb, 255:red, 255; green, 255; blue, 255 }  ,fill opacity=0.6 ] (128,69.57) .. controls (128,64.05) and (132.48,59.57) .. (138,59.57) .. controls (143.52,59.57) and (148,64.05) .. (148,69.57) .. controls (148,75.09) and (143.52,79.57) .. (138,79.57) .. controls (132.48,79.57) and (128,75.09) .. (128,69.57) -- cycle ;
\draw  [fill={rgb, 255:red, 255; green, 255; blue, 255 }  ,fill opacity=0.6 ] (128,169.57) .. controls (128,164.05) and (132.48,159.57) .. (138,159.57) .. controls (143.52,159.57) and (148,164.05) .. (148,169.57) .. controls (148,175.09) and (143.52,179.57) .. (138,179.57) .. controls (132.48,179.57) and (128,175.09) .. (128,169.57) -- cycle ;
\draw  [fill={rgb, 255:red, 255; green, 255; blue, 255 }  ,fill opacity=0.6 ] (128,144.57) .. controls (128,139.05) and (132.48,134.57) .. (138,134.57) .. controls (143.52,134.57) and (148,139.05) .. (148,144.57) .. controls (148,150.09) and (143.52,154.57) .. (138,154.57) .. controls (132.48,154.57) and (128,150.09) .. (128,144.57) -- cycle ;
\draw  [fill={rgb, 255:red, 255; green, 255; blue, 255 }  ,fill opacity=0.6 ] (128,94.57) .. controls (128,89.05) and (132.48,84.57) .. (138,84.57) .. controls (143.52,84.57) and (148,89.05) .. (148,94.57) .. controls (148,100.09) and (143.52,104.57) .. (138,104.57) .. controls (132.48,104.57) and (128,100.09) .. (128,94.57) -- cycle ;
\draw  [fill={rgb, 255:red, 74; green, 144; blue, 226 }  ,fill opacity=0.5 ] (198,40) -- (228,40) -- (228,180) -- (198,180) -- cycle ;
\draw  [fill={rgb, 255:red, 255; green, 255; blue, 255 }  ,fill opacity=0.6 ] (203,60) .. controls (203,54.48) and (207.48,50) .. (213,50) .. controls (218.52,50) and (223,54.48) .. (223,60) .. controls (223,65.52) and (218.52,70) .. (213,70) .. controls (207.48,70) and (203,65.52) .. (203,60) -- cycle ;
\draw  [fill={rgb, 255:red, 255; green, 255; blue, 255 }  ,fill opacity=0.6 ] (203,88) .. controls (203,82.48) and (207.48,78) .. (213,78) .. controls (218.52,78) and (223,82.48) .. (223,88) .. controls (223,93.52) and (218.52,98) .. (213,98) .. controls (207.48,98) and (203,93.52) .. (203,88) -- cycle ;
\draw  [fill={rgb, 255:red, 255; green, 255; blue, 255 }  ,fill opacity=0.6 ] (203,135) .. controls (203,129.48) and (207.48,125) .. (213,125) .. controls (218.52,125) and (223,129.48) .. (223,135) .. controls (223,140.52) and (218.52,145) .. (213,145) .. controls (207.48,145) and (203,140.52) .. (203,135) -- cycle ;
\draw  [fill={rgb, 255:red, 255; green, 255; blue, 255 }  ,fill opacity=0.6 ] (203,163) .. controls (203,157.48) and (207.48,153) .. (213,153) .. controls (218.52,153) and (223,157.48) .. (223,163) .. controls (223,168.52) and (218.52,173) .. (213,173) .. controls (207.48,173) and (203,168.52) .. (203,163) -- cycle ;
\draw  [fill={rgb, 255:red, 74; green, 144; blue, 226 }  ,fill opacity=0.5 ] (270,50) -- (300,50) -- (300,160) -- (270,160) -- cycle ;
\draw  [fill={rgb, 255:red, 255; green, 255; blue, 255 }  ,fill opacity=0.6 ] (275,65) .. controls (275,59.48) and (279.48,55) .. (285,55) .. controls (290.52,55) and (295,59.48) .. (295,65) .. controls (295,70.52) and (290.52,75) .. (285,75) .. controls (279.48,75) and (275,70.52) .. (275,65) -- cycle ;
\draw  [fill={rgb, 255:red, 255; green, 255; blue, 255 }  ,fill opacity=0.6 ] (275,90) .. controls (275,84.48) and (279.48,80) .. (285,80) .. controls (290.52,80) and (295,84.48) .. (295,90) .. controls (295,95.52) and (290.52,100) .. (285,100) .. controls (279.48,100) and (275,95.52) .. (275,90) -- cycle ;
\draw  [fill={rgb, 255:red, 255; green, 255; blue, 255 }  ,fill opacity=0.6 ] (275,141) .. controls (275,135.48) and (279.48,131) .. (285,131) .. controls (290.52,131) and (295,135.48) .. (295,141) .. controls (295,146.52) and (290.52,151) .. (285,151) .. controls (279.48,151) and (275,146.52) .. (275,141) -- cycle ;
\draw  [fill={rgb, 255:red, 74; green, 144; blue, 226 }  ,fill opacity=0.5 ] (342,62) -- (372,62) -- (372,152) -- (342,152) -- cycle ;
\draw  [fill={rgb, 255:red, 255; green, 255; blue, 255 }  ,fill opacity=0.6 ] (347,76) .. controls (347,70.48) and (351.48,66) .. (357,66) .. controls (362.52,66) and (367,70.48) .. (367,76) .. controls (367,81.52) and (362.52,86) .. (357,86) .. controls (351.48,86) and (347,81.52) .. (347,76) -- cycle ;
\draw  [fill={rgb, 255:red, 255; green, 255; blue, 255 }  ,fill opacity=0.6 ] (347,101) .. controls (347,95.48) and (351.48,91) .. (357,91) .. controls (362.52,91) and (367,95.48) .. (367,101) .. controls (367,106.52) and (362.52,111) .. (357,111) .. controls (351.48,111) and (347,106.52) .. (347,101) -- cycle ;
\draw  [fill={rgb, 255:red, 255; green, 255; blue, 255 }  ,fill opacity=0.6 ] (347,138) .. controls (347,132.48) and (351.48,128) .. (357,128) .. controls (362.52,128) and (367,132.48) .. (367,138) .. controls (367,143.52) and (362.52,148) .. (357,148) .. controls (351.48,148) and (347,143.52) .. (347,138) -- cycle ;
\draw  [fill={rgb, 255:red, 74; green, 144; blue, 226 }  ,fill opacity=0.5 ] (415,68) -- (445,68) -- (445,140) -- (415,140) -- cycle ;
\draw  [fill={rgb, 255:red, 255; green, 255; blue, 255 }  ,fill opacity=0.6 ] (420,104) .. controls (420,98.48) and (424.48,94) .. (430,94) .. controls (435.52,94) and (440,98.48) .. (440,104) .. controls (440,109.52) and (435.52,114) .. (430,114) .. controls (424.48,114) and (420,109.52) .. (420,104) -- cycle ;

\draw  [fill={rgb, 255:red, 255; green, 255; blue, 255 }  ,fill opacity=0.6 ] (420,127) .. controls (420,121.48) and (424.48,117) .. (430,117) .. controls (435.52,117) and (440,121.48) .. (440,127) .. controls (440,132.52) and (435.52,137) .. (430,137) .. controls (424.48,137) and (420,132.52) .. (420,127) -- cycle ;

\draw  [fill={rgb, 255:red, 255; green, 255; blue, 255 }  ,fill opacity=0.6 ] (420,80) .. controls (420,74.48) and (424.48,70) .. (430,70) .. controls (435.52,70) and (440,74.48) .. (440,80) .. controls (440,85.52) and (435.52,90) .. (430,90) .. controls (424.48,90) and (420,85.52) .. (420,80) -- cycle ;

\draw [color={rgb, 255:red, 0; green, 0; blue, 0 }  ,draw opacity=0.8 ][fill={rgb, 255:red, 128; green, 128; blue, 128 }  ,fill opacity=0.8 ]   (148,44.57) -- (203,60) ;
\draw [color={rgb, 255:red, 0; green, 0; blue, 0 }  ,draw opacity=0.8 ][fill={rgb, 255:red, 128; green, 128; blue, 128 }  ,fill opacity=0.8 ]   (148,69.57) -- (203,60) ;
\draw [color={rgb, 255:red, 0; green, 0; blue, 0 }  ,draw opacity=0.8 ][fill={rgb, 255:red, 128; green, 128; blue, 128 }  ,fill opacity=0.8 ]   (148,44.57) -- (203,135) ;
\draw [color={rgb, 255:red, 0; green, 0; blue, 0 }  ,draw opacity=0.8 ][fill={rgb, 255:red, 128; green, 128; blue, 128 }  ,fill opacity=0.8 ]   (148,94.57) -- (203,88) ;
\draw [color={rgb, 255:red, 0; green, 0; blue, 0 }  ,draw opacity=0.8 ][fill={rgb, 255:red, 128; green, 128; blue, 128 }  ,fill opacity=0.8 ]   (148,144.57) -- (203,135) ;
\draw [color={rgb, 255:red, 0; green, 0; blue, 0 }  ,draw opacity=0.8 ][fill={rgb, 255:red, 128; green, 128; blue, 128 }  ,fill opacity=0.8 ]   (148,169.57) -- (203,163) ;
\draw [color={rgb, 255:red, 0; green, 0; blue, 0 }  ,draw opacity=0.8 ][fill={rgb, 255:red, 128; green, 128; blue, 128 }  ,fill opacity=0.8 ]   (148,144.57) -- (203,163) ;
\draw [color={rgb, 255:red, 0; green, 0; blue, 0 }  ,draw opacity=0.8 ][fill={rgb, 255:red, 128; green, 128; blue, 128 }  ,fill opacity=0.8 ]   (148,169.57) -- (203,88) ;
\draw [fill={rgb, 255:red, 128; green, 128; blue, 128 }  ,fill opacity=0.8 ]   (148,144.57) -- (203,60) ;
\draw [fill={rgb, 255:red, 128; green, 128; blue, 128 }  ,fill opacity=0.8 ]   (148,69.57) -- (203,163) ;
\draw [color={rgb, 255:red, 0; green, 0; blue, 0 }  ,draw opacity=0.8 ][fill={rgb, 255:red, 128; green, 128; blue, 128 }  ,fill opacity=0.8 ]   (223,163) -- (275,90) ;
\draw [color={rgb, 255:red, 0; green, 0; blue, 0 }  ,draw opacity=0.8 ][fill={rgb, 255:red, 128; green, 128; blue, 128 }  ,fill opacity=0.8 ]   (223,88) -- (275,90) ;
\draw [color={rgb, 255:red, 0; green, 0; blue, 0 }  ,draw opacity=0.8 ][fill={rgb, 255:red, 128; green, 128; blue, 128 }  ,fill opacity=0.8 ]   (223,60) -- (275,90) ;
\draw [color={rgb, 255:red, 0; green, 0; blue, 0 }  ,draw opacity=0.8 ][fill={rgb, 255:red, 128; green, 128; blue, 128 }  ,fill opacity=0.8 ]   (223,163) -- (275,141) ;
\draw [color={rgb, 255:red, 0; green, 0; blue, 0 }  ,draw opacity=0.8 ][fill={rgb, 255:red, 128; green, 128; blue, 128 }  ,fill opacity=0.8 ]   (223,60) -- (275,141) ;
\draw [color={rgb, 255:red, 0; green, 0; blue, 0 }  ,draw opacity=0.8 ][fill={rgb, 255:red, 128; green, 128; blue, 128 }  ,fill opacity=0.8 ]   (223,135) -- (275,141) ;
\draw [color={rgb, 255:red, 0; green, 0; blue, 0 }  ,draw opacity=0.8 ][fill={rgb, 255:red, 128; green, 128; blue, 128 }  ,fill opacity=0.8 ]   (223,135) -- (275,65) ;
\draw [color={rgb, 255:red, 0; green, 0; blue, 0 }  ,draw opacity=0.8 ][fill={rgb, 255:red, 128; green, 128; blue, 128 }  ,fill opacity=0.8 ]   (223,88) -- (275,65) ;
\draw [color={rgb, 255:red, 0; green, 0; blue, 0 }  ,draw opacity=0.8 ][fill={rgb, 255:red, 128; green, 128; blue, 128 }  ,fill opacity=0.8 ]   (295,65) -- (347,101) ;
\draw [color={rgb, 255:red, 0; green, 0; blue, 0 }  ,draw opacity=0.8 ][fill={rgb, 255:red, 128; green, 128; blue, 128 }  ,fill opacity=0.8 ]   (295,141) -- (347,76) ;
\draw [color={rgb, 255:red, 0; green, 0; blue, 0 }  ,draw opacity=0.8 ][fill={rgb, 255:red, 128; green, 128; blue, 128 }  ,fill opacity=0.8 ]   (295,90) -- (347,76) ;
\draw [color={rgb, 255:red, 0; green, 0; blue, 0 }  ,draw opacity=0.8 ][fill={rgb, 255:red, 128; green, 128; blue, 128 }  ,fill opacity=0.8 ]   (295,65) -- (347,138) ;
\draw [color={rgb, 255:red, 0; green, 0; blue, 0 }  ,draw opacity=0.8 ][fill={rgb, 255:red, 128; green, 128; blue, 128 }  ,fill opacity=0.8 ]   (295,141) -- (347,101) ;
\draw [color={rgb, 255:red, 0; green, 0; blue, 0 }  ,draw opacity=0.8 ][fill={rgb, 255:red, 128; green, 128; blue, 128 }  ,fill opacity=0.8 ]   (367,76) -- (420,104) ;
\draw [color={rgb, 255:red, 0; green, 0; blue, 0 }  ,draw opacity=0.8 ][fill={rgb, 255:red, 128; green, 128; blue, 128 }  ,fill opacity=0.8 ]   (367,138) -- (420,104) ;
\draw [color={rgb, 255:red, 0; green, 0; blue, 0 }  ,draw opacity=0.8 ][fill={rgb, 255:red, 128; green, 128; blue, 128 }  ,fill opacity=0.8 ]   (367,138) -- (420,127) ;
\draw [color={rgb, 255:red, 0; green, 0; blue, 0 }  ,draw opacity=0.8 ][fill={rgb, 255:red, 128; green, 128; blue, 128 }  ,fill opacity=0.8 ]   (367,76) -- (420,127) ;
\draw [color={rgb, 255:red, 0; green, 0; blue, 0 }  ,draw opacity=0.8 ][fill={rgb, 255:red, 128; green, 128; blue, 128 }  ,fill opacity=0.8 ]   (367,101) -- (420,80) ;
\draw [color={rgb, 255:red, 0; green, 0; blue, 0 }  ,draw opacity=0.8 ][fill={rgb, 255:red, 128; green, 128; blue, 128 }  ,fill opacity=0.8 ]   (367,101) -- (420,127) ;
\draw   (501,88) -- (601,88) -- (601,118) -- (501,118) -- cycle ;
\draw   (624,100.5) -- (642,100.5) -- (642,98) -- (654,103) -- (642,108) -- (642,105.5) -- (624,105.5) -- cycle ;
\draw [fill={rgb, 255:red, 128; green, 128; blue, 128 }  ,fill opacity=0.8 ]   (295,90) -- (347,101) ;
\draw [color={rgb, 255:red, 0; green, 0; blue, 0 }  ,draw opacity=0.8 ][fill={rgb, 255:red, 128; green, 128; blue, 128 }  ,fill opacity=0.8 ]   (148,94.57) -- (203,60) ;
\draw [color={rgb, 255:red, 0; green, 0; blue, 0 }  ,draw opacity=0.8 ][fill={rgb, 255:red, 128; green, 128; blue, 128 }  ,fill opacity=0.8 ]   (295,141) -- (347,138) ;
\draw   (461,101) -- (479,101) -- (479,98.5) -- (491,103.5) -- (479,108.5) -- (479,106) -- (461,106) -- cycle ;
\draw [color={rgb, 255:red, 0; green, 0; blue, 0 }  ,draw opacity=0.8 ][fill={rgb, 255:red, 128; green, 128; blue, 128 }  ,fill opacity=0.8 ]   (148,69.57) -- (203,88) ;
\draw [color={rgb, 255:red, 0; green, 0; blue, 0 }  ,draw opacity=0.8 ][fill={rgb, 255:red, 128; green, 128; blue, 128 }  ,fill opacity=0.8 ]   (367,101) -- (420,104) ;
\draw  [line width=0.75]  (75,100.5) -- (93,100.5) -- (93,98) -- (105,103) -- (93,108) -- (93,105.5) -- (75,105.5) -- cycle ;
\draw  [dash pattern={on 4.5pt off 4.5pt}] (116.52,10) -- (610.52,10) -- (610.52,200) -- (116.52,200) -- cycle ;
\draw  [fill={rgb, 255:red, 109; green, 151; blue, 67 }  ,fill opacity=0.5 ] (43,47) -- (58,47) -- (58,159) -- (43,159) -- cycle ;

\draw (125,16) node [anchor=north west][inner sep=0.75pt]   [align=left] {{\scriptsize 4096x1}};
\draw (161,177) node [anchor=north west][inner sep=0.75pt]   [align=left] {{\tiny L-ReLU}};
\draw (139,111) node [anchor=north west][inner sep=0.75pt]  [rotate=-90] [align=left] {...};
\draw (215,106) node [anchor=north west][inner sep=0.75pt]  [rotate=-90] [align=left] {...};
\draw (288,108) node [anchor=north west][inner sep=0.75pt]  [rotate=-90] [align=left] {...};
\draw (238,156) node [anchor=north west][inner sep=0.75pt]   [align=left] {{\tiny L-ReLU}};
\draw (311,147) node [anchor=north west][inner sep=0.75pt]   [align=left] {{\tiny L-ReLU}};
\draw (382,143) node [anchor=north west][inner sep=0.75pt]   [align=left] {{\tiny L-ReLU}};
\draw (203,27) node [anchor=north west][inner sep=0.75pt]   [align=left] {{\scriptsize 512x1}};
\draw (275,37) node [anchor=north west][inner sep=0.75pt]   [align=left] {{\scriptsize 128x1}};
\draw (351,48) node [anchor=north west][inner sep=0.75pt]   [align=left] {{\scriptsize 64x1}};
\draw (424,56) node [anchor=north west][inner sep=0.75pt]   [align=left] {{\scriptsize 3x1}};
\draw (462,82) node [anchor=north west][inner sep=0.75pt]   [align=left] {{\tiny sigmoid}};
\draw (425,73.4) node [anchor=north west][inner sep=0.75pt]  [font=\scriptsize]  {$c_{x}$};
\draw (425,97.4) node [anchor=north west][inner sep=0.75pt]  [font=\scriptsize]  {$c_{y}$};
\draw (425,122.4) node [anchor=north west][inner sep=0.75pt]  [font=\scriptsize]  {$w$};
\draw (518,97) node [anchor=north west][inner sep=0.75pt]  [font=\footnotesize] [align=left] {Transform};
\draw (662,91.4) node [anchor=north west][inner sep=0.75pt]    {$( x_{tl} ,y_{tl} ,x_{br} ,y_{br})$};
\draw (28,174) node [anchor=north west][inner sep=0.75pt]   [align=center] {Feature\\Vector};
\draw (340,213) node [anchor=north west][inner sep=0.75pt]   [align=center] {
Regression\\
Head};

\end{tikzpicture}

\caption{Aspect Ratio Enforced Regression Head, where the transform generates the bounding box in the correct aspect ratio.}
\label{fig:regression_head}
\end{subfigure}
\caption{Proposed Model Framework}
\end{figure*}
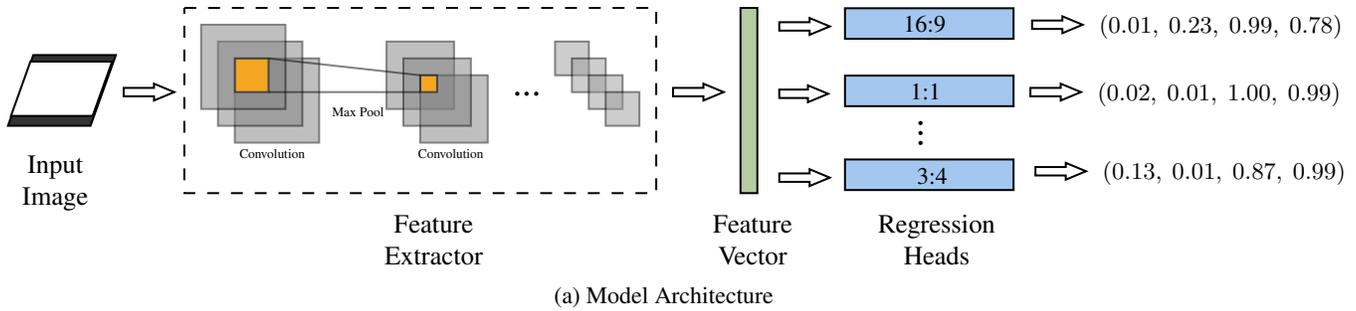
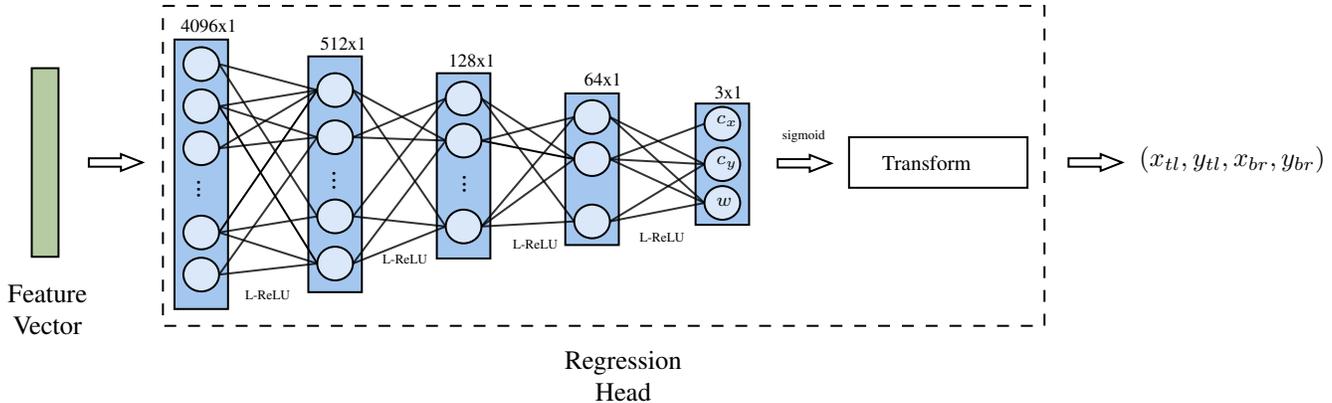

\subsection{Pre-processing and Augmentation}

We resize our images to $(224,224)$ and pad with zeros where necessary, to retain the aspect ratio of the original image. 
We store the bounds locations and the bounding boxes annotated by the editors as normalized coordinates of the top left and the bottom right corners. We then augment the image by randomly applying horizontal flips or color transformations such as changing the brightness or saturation or converting to grayscale. We do not apply any spatial transformation such as rotation or vertical flipping because these affect the composition of the image \cite{zeng2019grid}.

\subsection{Model}

\begin{figure}[t]
    \centering
    \includegraphics[width=\linewidth]{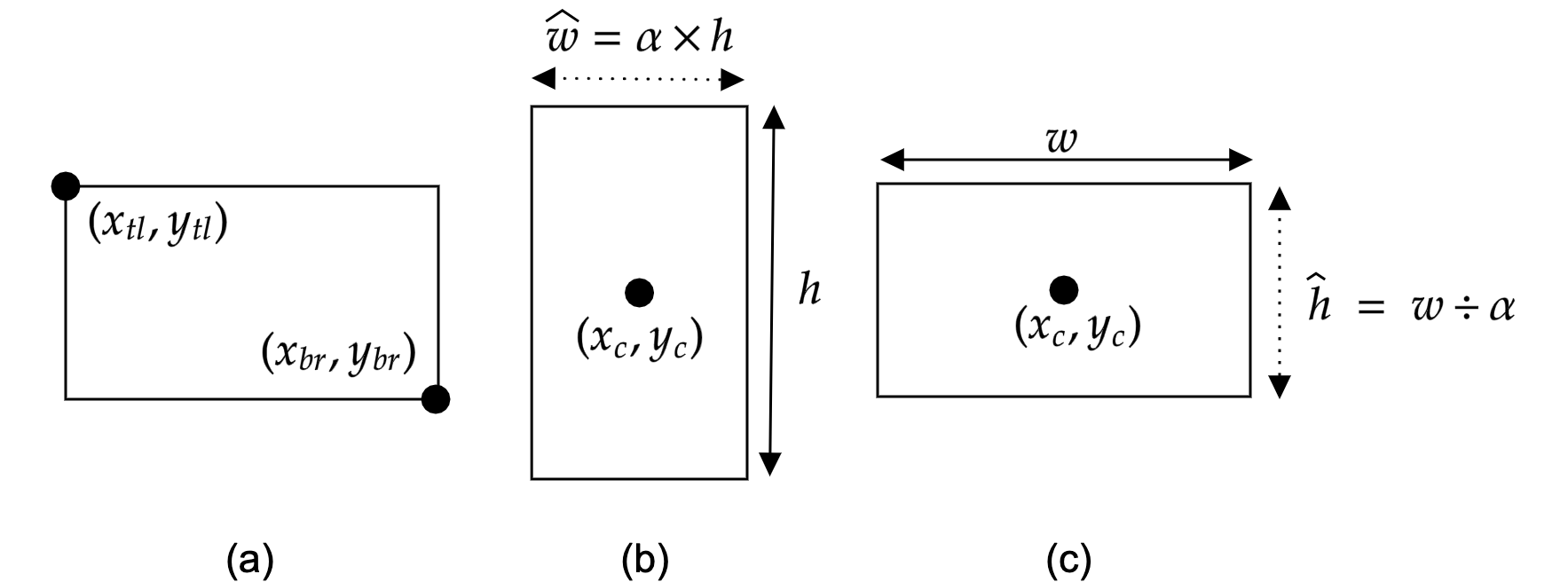}
    \caption{(a) Non-enforced bounding box prediction. (b) Enforced bounding box prediction where width $\hat{w}$ is inferred from the aspect ratio $\alpha$ (c) Enforced bounding box prediction where height $\hat{h}$ is inferred from the aspect ratio $\alpha$.}
    \label{fig:enforced_all_shrunk}
\end{figure}

Our proposed model can be conceptually divided into two modules - a shared \acrshort{cnn}-based feature extractor as the backbone, and multiple parallel regression heads, one for each aspect ratio. We illustrate this in Figure \ref{fig:model_architecture}. This design allows us to add predictor heads for new aspect ratios without having to retrain the rest of the network from scratch, or significantly increasing the time for inference. As illustrated in \ref{sec:mir_thumb_eval}, we are also able to generate crops for unseen aspect ratios without pre-training by leveraging the predictions from similar aspect ratios, which is helpful when training data is scarce.

\subsubsection{Feature Extractor} The feature extractor is designed to output a fixed-length feature vector for each input image, which is subsequently fed to the regression heads. We use a shared feature extractor because the regression head for each aspect ratio needs similar information to make a prediction, such as the important regions and their locations in the image. We try a few common \acrshort{cnn} architectures for the backbone, including VGG \cite{simonyan2014very}, ResNet \cite{he2016deep}, DenseNet \cite{huang2017densely}, WideResNet \cite{zagoruyko2016wide} and MobileNet-v2 \cite{sandler2018mobilenetv2}. These architectures have been used for many computer vision tasks, including some previous solutions to automatic image cropping \cite{lu2019listwise, tu2019image}. Another advantage of using common architectures is the wide availability of pre-trained network weights on Image Classification and related tasks. We study the effect of transfer learning using these pre-trained networks in Section \ref{sec:evaluation}.

\subsubsection{Regression Head} As shown in Figure \ref{fig:regression_head}, each regression head is a densely connected neural network, with Leaky ReLU as the activation function for the intermediate layers, and sigmoid activation at the output. Each regression head is dedicated to predicting a bounding box of a single aspect ratio, often represented as coordinates of the top-left $(x_{tl}, y_{tl})$  and the bottom-right corners $(x_{br},y_{br})$. However, predicting the bounding box using this representation does not guarantee that the output would correspond to the desired fixed aspect ratio.

We use an alternate regression head to predict images with a fixed aspect ratio, which we call an aspect ratio enforced regression head. For a landscape or square aspect ratio, we predict the coordinates of the center $(x_c,y_c)$ and the width $w$, using the aspect ratio $\alpha$ to predict the height. For a portrait aspect ratio, we predict the center coordinates $(x_c,y_c)$ and the height $h$, using $\alpha$ to predict the width. We illustrate this in Figure \ref{fig:enforced_all_shrunk}. Since the aspect ratio, $\alpha$, is fixed for a given regression head, we can draw a bounding box by calculating the remaining dimension, represented by the transform operation in Figure \ref{fig:regression_head}. At run time, we clip the prediction bounding box to the largest possible bounding box for the given image and the predicted center coordinates ($x_c,y_c$) to avoid invalid output. We use the Smooth L1 loss between the annotated and the predicted bounding box coordinates, similar to Fast R-CNN \cite{girshick2015fast}. Our experiments below were performed with models with a single enforced regression head per aspect ratio. The architecture could be extended to include multiple regression heads per aspect ratio if desired. This could be useful in cases where, for example, a close-up version and a zoomed out version for each aspect ratio are needed.

\section{Experiments and Ablation Study}
\label{sec:experiments}
We perform a 60/20/20 split on our dataset to create training, validation, and test sets. We use the ADAM \cite{kingma2014method} optimizer with a default learning rate of $0.0001$ and a batch size of 128, and use early stopping with the validation set.

We use the Boundary Displacement Error (BDE) and the Intersection over Union (IoU) to evaluate cropping, in line with previous approaches \cite{lu2019end,lu2019aesthetic,tu2019image,guo2018automatic,celona2019autocropping}. As we did not have editors rank or rate different crops, we cannot compute ranking metrics or leverage ranked learning approaches. All metrics in the tables in this section are averaged across all the aspect ratios in our test set.









\begin{table*}[t]
\centering
\begin{tabular}{lccccccc}
\hline
\textbf{Model} & \textbf{Pre-Training} & \textbf{Train Set} & \textbf{Enforced} & \textbf{IoU} $\uparrow$ & \textbf{BDE} $\downarrow$ \\ \hline
Baseline-0.8 & - & - & - & 0.645 & 0.076 \\ 
Baseline-0.9 & - & - & - & 0.697 & 0.061 \\
Baseline-1.0 & - & - & - & \textbf{0.728} & \textbf{0.053} \\ \hline
GAIC\cite{zeng2019grid} & ImageNet & GAIC\cite{zeng2019grid} & True* & 0.723 & 0.058 \\ \hline
Ours (WideResNet-50-2) & ImageNet & Ours & True & \textbf{0.867} & \textbf{0.023} \\
Ours (WideResNet-50-2) & ImageNet & Ours & False & 0.855 & 0.025 \\
Ours (WideResNet-50-2) & None & Ours & True & 0.832 & 0.030 \\ \hline

\end{tabular}
\caption{Model evaluation on our dataset.}
\label{table:eval_all}
\end{table*}

\subsection{Evaluation on our Dataset}\label{sec:evaluation}

We compare our models with a baseline method that predicts bounding boxes of varying sizes in the correct aspect ratio around the center of the image \cite{celona2019autocropping}. We denote this family of methods, \emph{Baseline-$s$}, where $s$ represents the scaling factor of the bounding box as a fraction of the largest possible bounding box for that aspect ratio. We also compare our method with GAIC\cite{zeng2019grid}, a recent method capable of cropping images in fixed aspect ratios. This is enabled by selecting the aspect ratio of the generated candidates, different from our enforced predictor head method which is more efficient. We use the trained models and code released by the authors, but are unable to fine-tune GAIC on our training set as GAIC relies on densely annotated images which are not available in our dataset. Our consolidated results, evaluated on our dataset can be seen in Table \ref{table:eval_all}.

We also study the influence of pre-training on the ImageNet dataset for the feature extractor component of the model. We find that pre-training offers significant performance improvements, and present our results in Table \ref{table:eval_all}, likely because both tasks require the model to learn the position and the type of objects in an image. We use transfer learning for all subsequent experiments.

\subsubsection{Enforced Aspect Ratio Prediction}
We test our method of enforcing the aspect ratio of the bounding box, and report the results with non-enforced and enforced predictions in Table \ref{table:eval_all}. The aspect ratio enforced prediction method improves model performance while also satisfying the exact aspect ratio requirement.

\subsubsection{CNN Backbone Architecture}
\label{sec:baseline} 
We experiment with various common CNN architectures pre-trained on ImageNet for the feature extractor. We use enforced aspect ratio regression heads, keep all other hyper-parameters such as learning rate constant and present the metrics in Table \ref{table:backbone_architecture}. We find that the WideResNet-50-2 architecture performs the best on our test set overall. We also find that MobileNet-v2 performs very well, considering its smaller size in terms of the number of trainable parameters.

\begin{table}[t]
\centering
\begin{tabular}{cccc}
\hline
\textbf{Model} & Size & IoU $\uparrow$ & BDE $\downarrow$ \\ \hline
VGG16 & 138.3M & 0.854 & 0.025 \\ \hline
WideResNet-50-2 & 68.8M & \textbf{0.867} & \textbf{0.023} \\ 
ResNet-50 & 25.5M & 0.861 & 0.025 \\ 
ResNeXt-50 & 25.0M & 0.846 & 0.027 \\ \hline
Densenet-121 & 7.9M & 0.860 & 0.025 \\ \hline
MobileNet-v2 & 3.5M & 0.854 & 0.025 \\ \hline
\end{tabular}
\caption{CNN Backbone Architecture Comparison, pre-trained on ImageNet and fine-tuned on our dataset}
\label{table:backbone_architecture}
\end{table}



\subsection{Evaluation on FCDB}
The datasets most commonly used to evaluate automatic image cropping methods like FCDB \cite{chen2017quantitative} do not impose any requirements on aspect ratios. Since our model is designed to predict fixed aspect ratios, this makes an accurate benchmark difficult. Nevertheless, we modify our model for this experiment to produce bounding boxes in any aspect ratio. Specifically, we remove the aspect ratio enforced regression heads and attach a single non-enforced regression head to the trained feature extractor. We further split the FCDB training set 80/20 into a training and validation split, and then fine-tune our modified model on the resulting training split using a batch size of 128, and an ADAM optimizer with a learning rate of $1 \times 10^{-5}$ for 300 epochs. We use early stopping on the validation split, similar to the previous experiments. 

We present metrics on the FCDB test set in Table \ref{table:evaluation_fcdb}. We report the metrics of VFN \cite{chen2017learning} and VPN \cite{wei2018good} as in Lu et al. \cite{lu2019listwise}, without including the ground truth window as a candidate view for VFN, and without the post processing step in VPN.

The results demonstrate that our approach is competitive with other models that explicitly model image aesthetics or visual attention regions without evaluating multiple crop candidates. Our model achieves a higher IoU score than the end-to-end model by Lu et al. \cite{lu2019end}, with a more straightforward training approach that does not require the identification of visual attention regions. LVRN \cite{lu2019listwise} achieves a slightly higher IoU score, but evaluates an average of 1,745 candidates per image, which is inefficient. The ASM-Net \cite{tu2019image} achieves a higher IoU score, but uses an inefficient two-stage searching step and derives composition patterns from human-defined composition rules that may not generalize. Out of these, VFN \mbox{\cite{chen2017learning}}, LVRN \mbox{\cite{lu2019listwise}}, Wang et al. \mbox{\cite{wang2018deep}} and Lu et al. \mbox{\cite{lu2019end}} perform fine-tuning on the FCDB training set, while the authors of the other approaches only use the FCDB test set for evaluation purposes. We are not able to find a significant influence of fine-tuning on the metrics, likely because of the relatively small size of FCDB (1395 train and 348 test images) and the wide differences between individual approaches. Our model's performance is comparable to or better than the subset of models that fine-tune on FCDB, and is significantly more efficient.


We also study the impact of transfer learning on our dataset, by initializing the feature extractor using the weights from ImageNet and training the model as described above. The resulting model labeled "Ours-ImageNet Only", achieves an IoU of $0.679$, slightly worse than the model initialized with the weights learned on our dataset, but better than Lu et al. \cite{lu2019end}, VFN \cite{chen2017learning} and VPN \cite{wei2018good}. This implies that the proposed architecture is a more significant contributor to the performance on FCDB, compared to pre-training on our dataset.




Prior approaches measured their efficiency during using the time to crop a single image as a metric, which is dependent on hardware, input image size and implementation details, making a fair comparison difficult. Nevertheless, we present the number of crops per second and hardware self-reported by the authors in Table \ref{table:evaluation_fcdb} as a measure of efficiency. Amongst the approaches compared, sliding-judging methods such as VFN \cite{chen2017learning} have the lowest efficiency. More recent approaches from Lu et al. \cite{lu2019end, lu2020learning} report an overall processing speed of 50fps for their image cropping solution an Nvidia 2080Ti GPU. The weakly supervised approach by Lu et al. \cite{lu2020weakly} is able to crop images at 285 fps.  
In contrast, our model with a WideResNet-50-2 backbone can crop 606 input frames per second on a single Nvidia Tesla V100 GPU in 5 different aspect ratios simultaneously, resulting in over 3000 output crops per second. This time is calculated for inference only without any optimizations, and does not include time to load and pre-process the images, or save the final cropped images in order to be consistent and enable comparisons with prior work\mbox{\cite{chen2018cropnet}}.



\begin{table*}[t]
\centering
\begin{tabular}{lccccccc}
\hline
\textbf{Model} & \textbf{Fine Tuning} & \textbf{Avg. Candidates} & \textbf{IoU} $\uparrow$ & \textbf{BDE} $\downarrow$ & \textbf{FPS}$\uparrow$ & \textbf{GPU Hardware} \\ \hline
VFN \cite{chen2017learning} & Yes & 137 & 0.632 & 0.098 & 0.78 \cite{lu2020weakly} & N/A \\\hline
A2-RL \cite{li2018a2} & No & 13.56 & 0.663 & 0.089 & 4.08 & Nvidia Titan X \\ \hline
VPN \cite{wei2018good} & No & 895 & 0.664   & 0.085 & 75 & N/A \\\hline
Wang et al.\text{*} \cite{wang2018deep} & Yes & 1296 & 0.65 & 0.08 & - & -  \\
\hline
LVRN \cite{lu2019listwise} & Yes & 1745 & 0.7100 & 0.0735 & 125 & Nvidia 1080  \\
\hline
ASM-Net \text{*} \cite{tu2019image} & No & N.A. & 0.748 & 0.068 & - & -  \\ \hline

Lu et al.\text{*} \cite{lu2019end} & Yes & 1 & 0.673 & 0.058 & 50 & Nvidia 2080 Ti  \\  \hline
Lu et al.\text{*} \cite{lu2020learning} & No & 1 & 0.673 & 0.058 & 50 & Nvidia 2080 Ti   \\ \hline
Lu et al.\text{*} \cite{lu2020weakly} & No & 1 & 0.681 & 0.084 & 285 & Nvidia 1080 Ti \\ \hline

\textbf{Ours-ImageNet Only} & Yes & 1 & 0.679 & 0.067 & 606 & Nvidia Tesla V100 \\ 
\textbf{Ours} & Yes & 1 & 0.692 & 0.064 & 606 & Nvidia Tesla V100 \\ 

\hline

\end{tabular}
\caption{Evaluation on FCDB, where \text{*} highlight models that explicitly model aesthetics and/or attention regions. FPS refers to the number of input frames per second. 
}
\label{table:evaluation_fcdb}
\end{table*}

\subsection{Evaluation on MIR-Thumb}
\label{sec:mir_thumb_eval}
Even though the goals of image cropping methods differ from those of thumbnail generation, datasets like MIR-Thumb used by CropNet \cite{chen2018cropnet} are similar to ours, in that they annotate the same image with bounding boxes of different aspect ratios. We, therefore, evaluate our model on the MIR-Thumb test set to test its generalization ability. We also include our baseline methods from Section \ref{sec:baseline} for comparison.

MIR-Thumb includes some aspect ratios that were not present in our dataset, namely 21:9, 9:16, and 9:21. We synthesize the model predictions for these aspect ratios by adjusting the bounding boxes of the closest aspect ratio in our model. To generate the target aspect ratio of 21:9, we reduce the height uniformly around the center of the 2:1 prediction in our model and keep the width constant. The closest aspect ratio to 9:16 and 9:21 was 3:4, where we keep the height constant and shrink the width. The results are shown in Table \ref{table:eval_mir_thumb}, where our model achieves state-of-the-art performance with no fine-tuning. We achieve a significantly higher IoU of 0.770 when we only consider aspect ratios that were in our training set namely 16:9, 1:1, 4:3, and 3:4. These results indicate that our learned model generalizes well to other similar datasets and tasks.

\begin{table}[t]
\centering
\begin{tabular}{lccc}
\hline
\textbf{Model} & \textbf{Train Set} & \textbf{Test Set} & \textbf{IoU} $\uparrow$   \\ \hline
Baseline-0.8 & None & MIR-Thumb & 0.488 \\ 
Baseline-0.9 & None & MIR-Thumb & 0.505 \\ 
Baseline-1.0 & None & MIR-Thumb & 0.506 \\ \hline
CropNet & FAT-Clean \cite{chen2018cropnet} & MIR-Thumb & 0.672 \\ 
CropNet & MIR-Thumb & MIR-Thumb & 0.711 \\ \hline
\textbf{Ours} & Ours & MIR-Thumb & \textbf{0.741} \\ \hline
\end{tabular}
\caption{Evaluation on MIR-Thumb}
\label{table:eval_mir_thumb}
\end{table}

\subsection{Qualitative Assessment}
\label{sec:qualitative}
\begin{figure}[ht]
    \centering
    \includegraphics[width=0.9\linewidth]{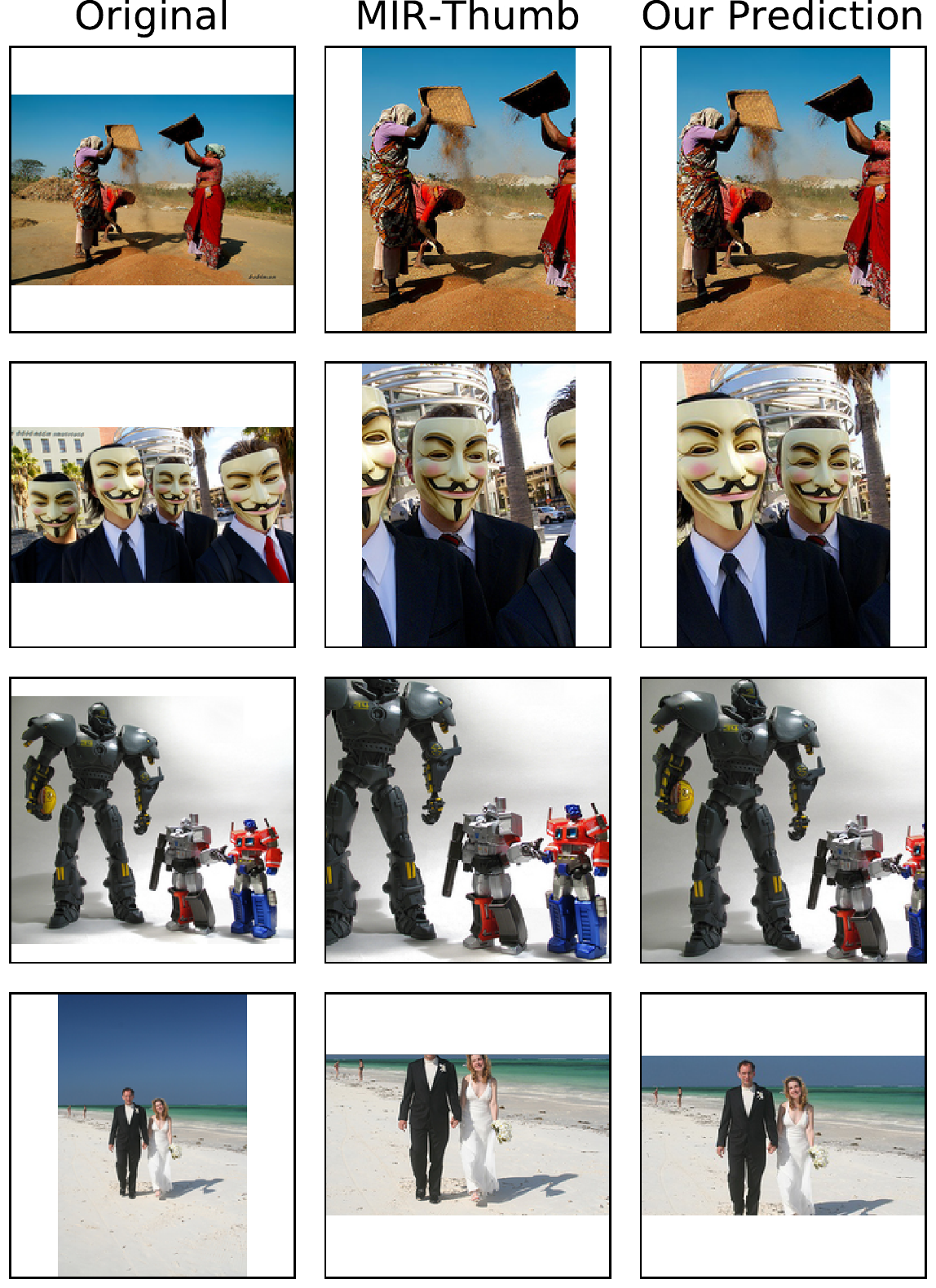}
    \caption{Our model can generate crops with better aesthetics than the original annotations in MIR-Thumb without fine-tuning}
    \label{fig:mir_thumb_improvements}
\end{figure}

The perception of image aesthetics is inherently subjective. We, therefore, provide visual examples of some results of our algorithm on images from MIR-Thumb, FCDB, and our dataset. We first illustrate a few cases where our model with no fine-tuning produced crops with better aesthetics than even the annotated images in the MIR-Thumb test set, as seen in Figure \ref{fig:mir_thumb_improvements}. Most of these cases involve partially cropped human subjects which are rare in our training dataset but appear more frequently in the MIR-Thumb dataset, resulting in our model predictions getting a low IoU score, even though the predicted crops have arguably better aesthetics. This finding reveals some open challenges in the objective evaluation of image cropping systems, such as using the IoU as a metric, the reliance on a single reference annotation and using inexperienced crowd-sourced workers. Future research in these areas is critical in order to build reliable and robust image cropping systems.

\begin{figure}[ht]
    \centering
    \includegraphics[width=0.9\linewidth]{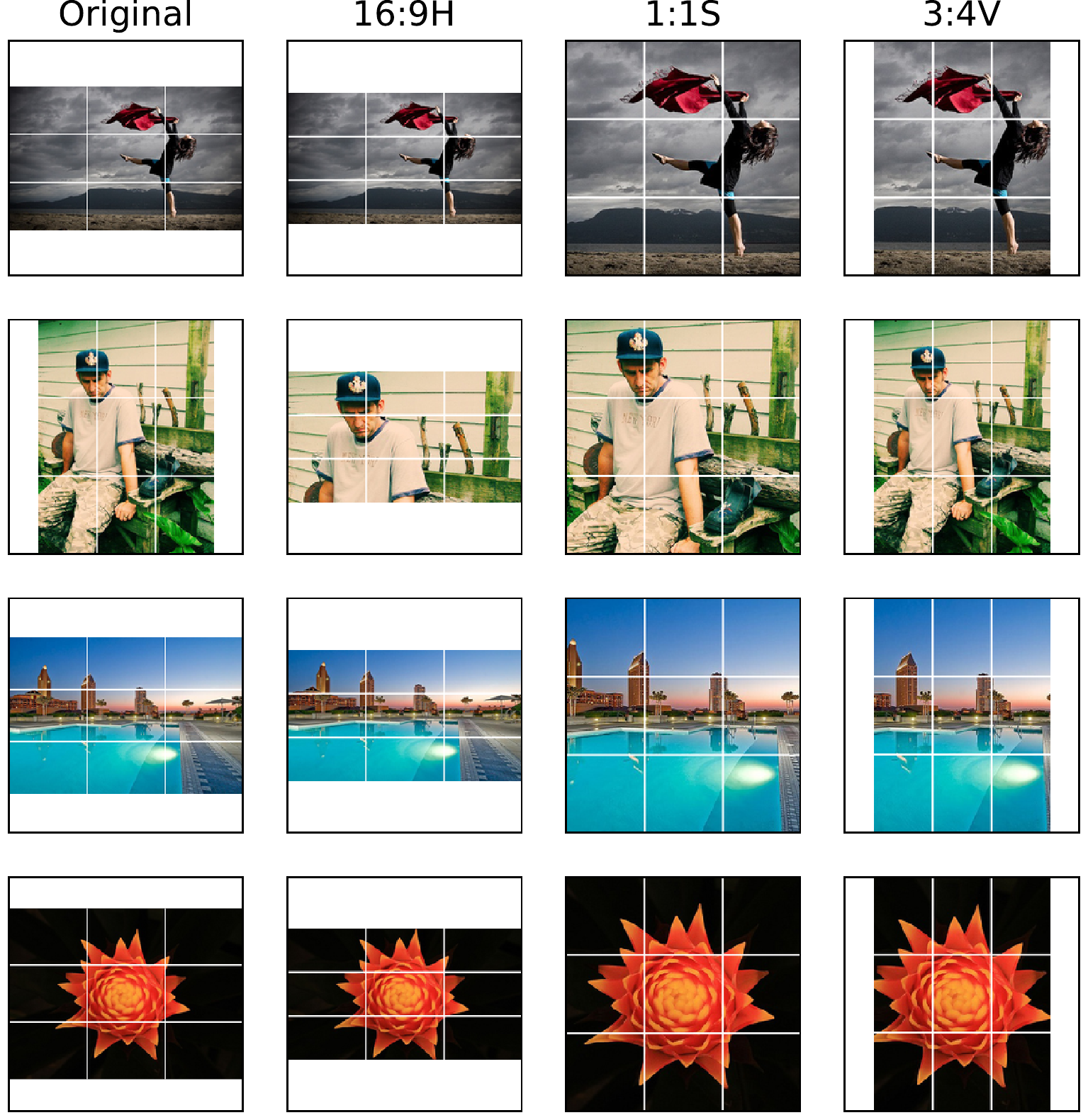}
    \caption{Illustrating the model's ability to retain aesthetic and composition properties (eg. rule of thirds) of the original image, evaluated on images sourced from the MIR-Thumb dataset without fine-tuning.}
    \label{fig:rule_of_thirds}
\end{figure}

Additionally, we find that our model predictions appear to retain some composition aspects from the original image without explicitly modeling aesthetics during training. To illustrate this, we draw a rule-of-thirds grid over some images from MIR-Thumb and the resulting predictions in Figure \ref{fig:rule_of_thirds}. This behavior is consistent even when the aspect ratios of the source images and the target crop are quite different. We believe this behavior is a likely result of our dataset that includes well-composed source images cropped by editorial experts, unlike other datasets for image cropping such as FLMS that exclude well-composed images, assuming that they do not require further cropping \cite{fang2014automatic}.





\begin{figure*}[h]
    \centering
    \includegraphics[width=0.9\textwidth,height=0.93\textheight,keepaspectratio]{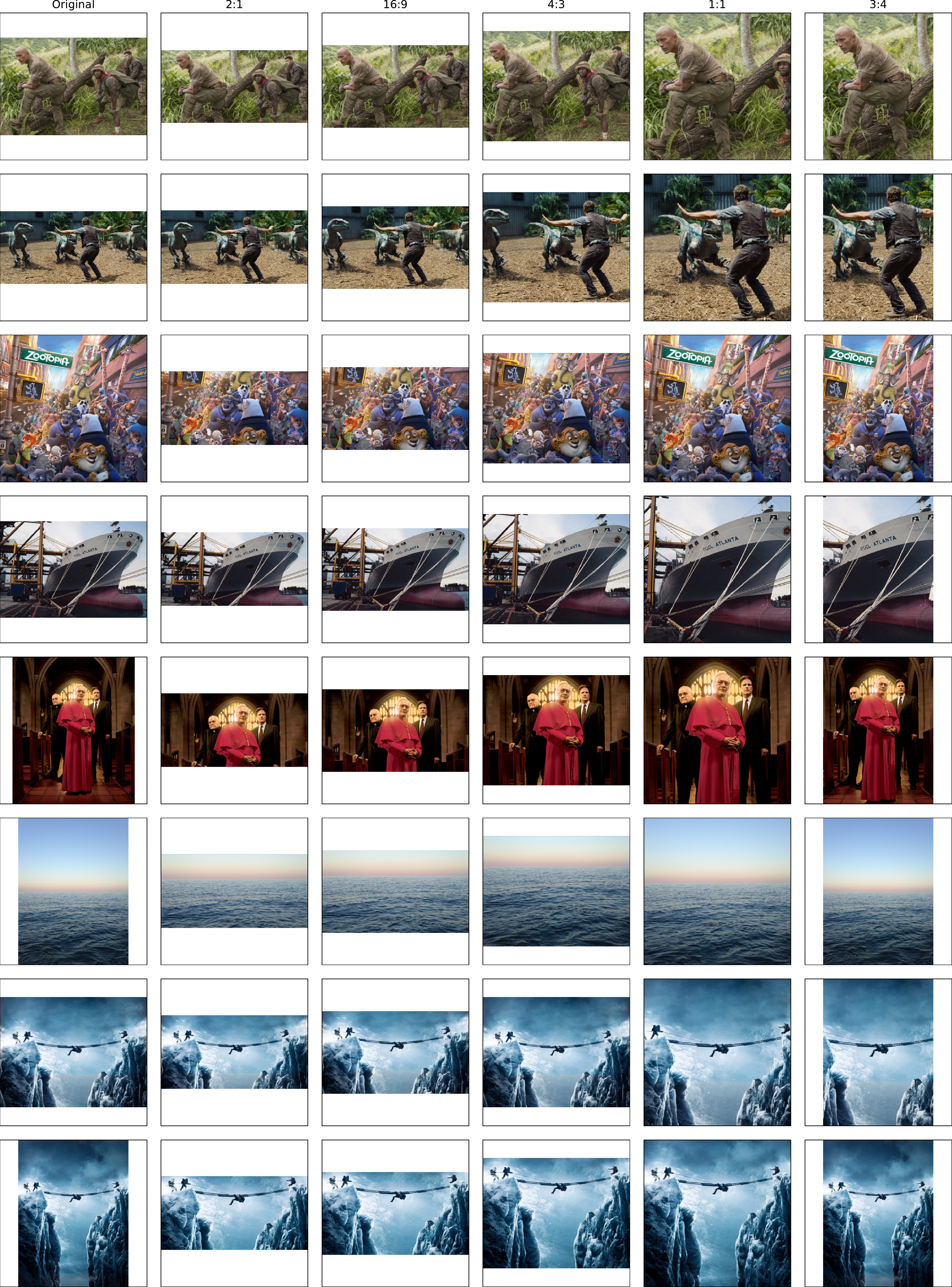}
    \caption{Results on images from our test dataset}
    \label{fig:results_montage}
\end{figure*}

Furthermore, we include some of the model predictions from our test set in Figure \ref{fig:results_montage}. The model can identify the main subject in the image, even if the subject is relatively small, facing away from the camera or is inanimate. 
The last two rows in Figure \ref{fig:results_montage} are intended to display predictions when we input two images with similar content but different aspect ratios. In both cases, the model can preserve the regions of interest while producing aesthetically similar crops for many of the output aspect ratios. 

\begin{figure*}[h]
    \centering
    \includegraphics[width=0.9\textwidth,height=0.93\textheight,keepaspectratio]{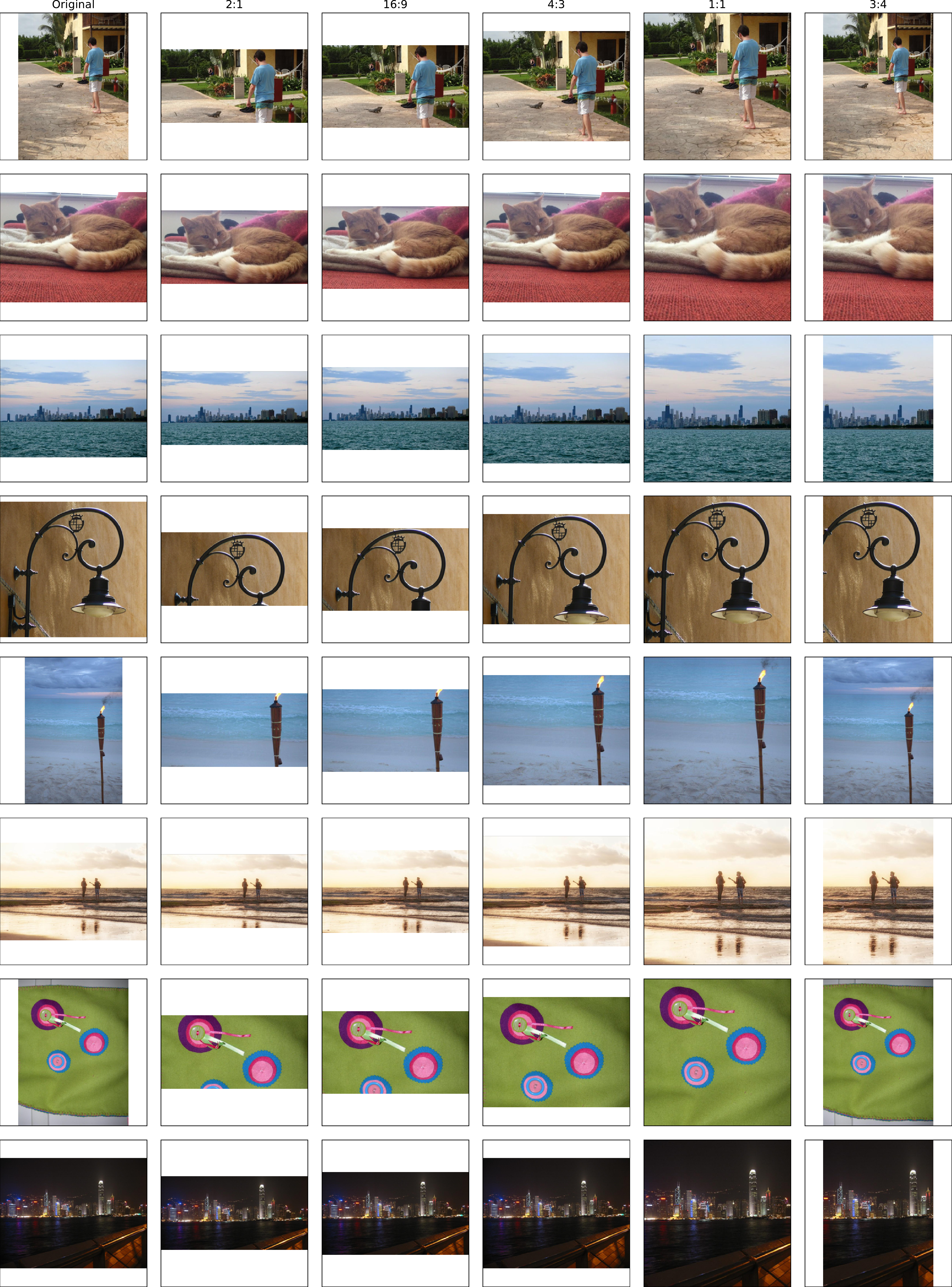}
    \caption{Results from our model trained on our dataset and evaluated on images in FCDB with no fine tuning}
    \label{fig:results_montage_fcdb}
\end{figure*}

We finally include some examples of the model predictions on FCDB without any fine tuning in Figure \mbox{\ref{fig:results_montage_fcdb}}, to illustrate the generalization ability of our model on a different dataset. The model is able to perform well on challenging and diverse images such as close up images of pets, day and night time landscapes, abstract patterns and inanimate objects. We observe that the model is able to retain important image content even in difficult cases when a large portion of the image has to be excluded, such as choosing a 2:1 crop of a portrait image (as seen in rows 1, 5, and 7 of Figure \mbox{\ref{fig:results_montage_fcdb}}).

\section{Conclusions}
We proposed a novel \emph{experience-based direct generation} strategy for image cropping. The model was designed to directly predict bounding boxes for a fixed aspect ratio, without explicitly modeling image aesthetics or visual attention regions. The model was trained on a large dataset of images annotated by experts, who tried to maintain image aesthetics and visual attention regions in the cropped images. We designed an efficient, straightforward architecture with a shared feature extractor and multiple dedicated regression heads to simultaneously predict the bounding box for different aspect ratios. Our model is easier to train than existing multi-stage approaches, and more efficient for inference as it does not evaluate multiple candidates.

Due to a lack of public datasets for our task, we benchmarked our model on two related datasets - FCDB for aesthetic image cropping without regard to aspect ratio, and MIR-Thumb for image thumbnail generation in fixed aspect ratios where aesthetics are not crucial. Our model, modified to generate outputs without defined aspect ratios, achieved results comparable to existing approaches, while being more efficient and easier to train. We achieved state-of-the-art results on the MIR-Thumb dataset without fine-tuning. Finally, we displayed some examples where our model generates more aesthetic crops than the ground truth annotations in MIRThumb. We also performed a qualitative evaluation and showed that our model is able to generalize across multiple datasets without fine-tuning, and also frequently retain aesthetic properties of the source image in the final crops.

{\small
\bibliographystyle{ieee_fullname}
\bibliography{egbib}
}

\end{document}